# Learning to Walk: Spike Based Reinforcement Learning for Hexapod Robot Central Pattern Generation

Ashwin Sanjay Lele, Yan Fang, Justin Ting, Arijit Raychowdhury
School of Electrical and Computer Engineering
Georgia Institute of Technology, Atlanta, GA, USA
{alele9, jting31, yan.fang}@gatech.edu, arijit.raychowdhury@ece.gatech.edu

**Learning to walk – i.e., learning locomotion under performance and energy constraints continues to be a challenge in legged robotics. Methods such as stochastic gradient, deep reinforcement learning (RL) have been explored for bipeds, quadrupeds and hexapods. These techniques are computationally intensive and often prohibitive for edge applications. These methods rely on complex sensors and pre-processing of data, which further increases energy and latency. Recent advances in spiking neural networks (SNNs) promise significant reduction in computing owing to the sparse firing of neuros and has been shown to integrate reinforcement learning mechanisms with biologically observed spike time dependent plasticity (STDP). However, training a legged robot to walk by learning the synchronization patterns of central pattern generators (CPG) in an SNN framework has not been shown. This can marry the efficiency of SNNs with synchronized locomotion of CPG based systems – providing breakthrough end-to-end learning in mobile robotics. In this paper, we propose a reinforcement based stochastic weight update technique for training a spiking CPG. The whole system is implemented on a lightweight raspberry pi platform with integrated sensors, thus opening up exciting new possibilities.**

## I. INTRODUCTION

Locomotion of animals emerge from the biomechanical interaction between the environment and the body controlled by the nervous systems [1]. Locomotion of invertebrate and vertebrates with limbs is generated by Central pattern generators (CPGs), which are biological neural circuits that produce rhythmic outputs [2][3]. CPG-inspired control systems have been widely applied to the locomotion of legged robots and exoskeletal prosthetic systems [4-6]. Compared to a traditional centralized control that calculates the activity of each leg independently, a distributed CPG-based de-centralized approach can locally control each joint through a combination of local and global network dynamics. Thus, it decreases the dimensionality of the control signal and reduce the time delays in motor control loop. Researchers also discovered that CPG not only gives rise to complex rhythmic patterns for locomotion and seamless gait transition but also receives modulation from higher regions in the brain. Inspired by these findings in neurosciences, we develop an end-to-end autonomous system where sensory inputs can directly interface with CPG to not only provide a means of adaptive locomotion and gait; but also enable the agent to walk or run without any prior knowledge or model-based experience replay.

Besides enabling functional end-to-end learning, we also consider the energy consumption of such autonomous systems. This is particularly important for micro-robots for real-world tasks in energy-constrained environments. Recent advances in spiking neural networks (SNN) and neuromorphic computing hardware has tremendously improved the energy-efficiency of cognitive tasks thus bridging the gap between hardware and wetware [7][8].

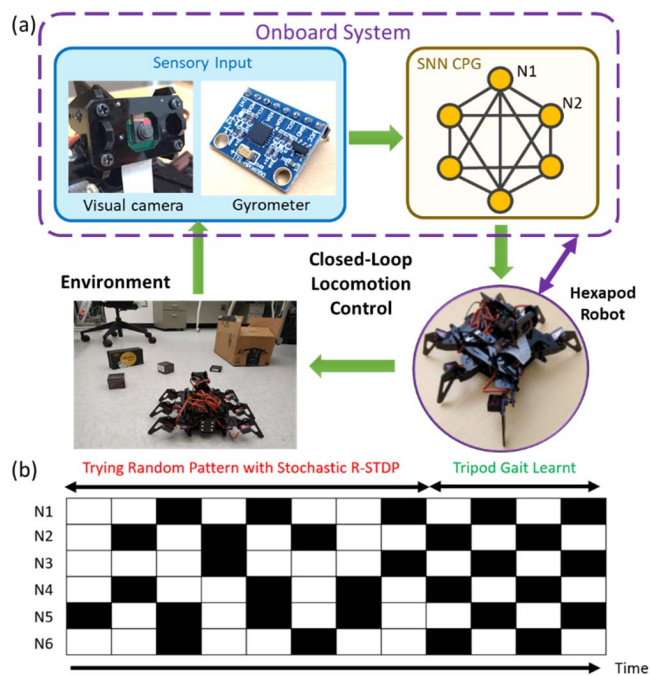

Fig. 1. (a) Closed loop locomotion system schematic. Hexapod robot moving in the environment sends visual and gyros sensor signals to the raspberry pi processing unit to decide which legs to move during the next time instance (b) Gait learning with time. Black boxes indicate that the neuron spikes at that time instance. Initially the CPG neurons spikes randomly making the robot to lose balance which is corrected over time with the robot learning the correct gait

In prior work, a CPG based on digital implementation of phase patterns has been proposed and implemented on an FPGA [5]. However, the generated patterns are pre-programmed, to enable known walking patterns. Evolutionary algorithm-based approaches are also proposed for gait learning of hexapod robots [9], but the neural CPG is not discussed or related to control. [10-11] show reinforcement learning formulation in the context of SNNs. However, most of these works focus on image processing or task planning of wheeled robots. To the best of our knowledge, SNN based CPG with autonomous learning ability has not been explored yet..

In this work, we demonstrate an end-to-end spiking neural network carrying out online processing of sensory information to generate gait learning in hexapod robots. Gyro sensor and camera provide the sensory inputs that are converted into reward signals reinforcing or depressing the weights to stochastically evolve the network to learn the tripod gait for locomotion. Thus we demonstrate a system where a legged robot learns to walk autonomously through interactions with the environment and by collecting data from its gyroscopic and camera sensors.

## II. PROPOSED SYSTEM

### A. Network Architecture

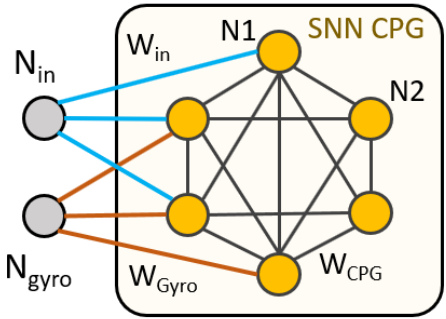

Fig. 2. Schematic of the neural connectivity within the network. The box forms the CPG driven by the input neuron and balance driven neuron. $W_{in}$, $W_{Gyro}$ and $W_{CPG}$ represent the weight matrices corresponding to the all-o-all connection. Only a few connections are shown in the figure due to space constraint

The proposed CPG network consists of six leaky-integrate and fire (LIF) spiking neurons, each driving one leg of the hexapod robot shown in Fig. 1. In this work, we used a digital discretized version of LIF model [5]. The update of variables occurs at discrete time steps $t$. The membrane voltage of LIF neurons take the injection current $I$ from spikes of other neuron and simultaneously decay with a leaky constant α. When the membrane potential $V$ reaches a spiking threshold, a spike is issued and $V$ is reset to the resting potential (zero in our case). Thus, we have the discrete equation of membrane voltage:

$$V[t+1] = \frac{V[t]}{\alpha} + I \quad (1)$$

The neurons are fully connected to each other with additive synapses. The CPG is driven by an input neuron with periodic input causing the CPG to fire and another neuron which fires only when the robots loses balance while walking. This neuron is a gyro sensor driven neuron referred as $N_{Gyro}$. The CPG neurons have a refractory period of two time units. The compactness of the network enables its implementation on the Raspberry Pi board without any external infrastructural support.

### B. Algorithm

Fig. 3 shows the pseudo-code for the algorithm. The weights are randomly initialized. Equations in step 7 and 8 show the voltage calculation for the LIF neurons where α is the leak rate.

The voltages of all the LIF neurons in the CPG are calculated by calculating the current coming into each neuron from the input neuron, gyro sensor driven neuron and other CPG neurons. The current is assumed to be equal to the weight of the synapse if the pre-neuron has spiked and zero otherwise as shown in the equation of step 7. An output neuron fires when the voltage exceeds a predefined spiking threshold. At any time instance, the number of neurons firing result in movement of the corresponding legs in the forward direction.

For every time instance, visual and gyro inputs are received before and after the movement is completed. During the movement of the leg, gyro sensor reading provides information about the balance and at the completion of the movement the visual information from the camera conveys whether a forward movement has occurred. If the balance is lost, the gyro driven neuron fires in the next time instance and the moved leg is taken back to the initial position. If the balance is not lost the movement of the leg is completed making the body to move forward. The balance and movement are then fed into the reward calculation equation. The detailed experimental design for the weight updating is explained in the next section.

**Algorithm 1** Learn to Walk
1: Initialize weights randomly $W_{in}, W_{Gyro}, W_{CPG}$
2: Initialize CPG neuron voltages, $V_{CPG}[0] = 0$
3: Initialize Spikes $S_{in} = 1, S_{Gyro} = 0, S_{CPG} = 0$
4: **for** $time = 1$ to $T$ **do**
5:    Read camera and gyro sensor for initial reading
6:    **for** $neuron = 1$ to $6$ **do**
7:      $I = W_{in}S_{in} + W_{Gyro}S_{Gyro} + W_{CPG}S_{CPG}$
8:      $V_{neuron}[t] = V_{neuron}[t-1]/\alpha + I$
9:      **if** ($V_{neuron}[t] > V_{Thresh}$) **then**
10:        Update Spike, $S_{CPG} = 1$, $V_{neuron}[t] = 0$
11:        Move Corresponding Leg
12:      **end if**
13:    **end for**
14:    Read camera and gyro sensor for final reading
15:    Calculate reward
16:    Update weights
17: **end for**

Fig. 3. Pseudo-code for the algorithm. For every time instance, the membrane voltages of neurons are updated to determine the neural spiking at the current time instance. After the corresponding legs are moved, the gyro sensor reading and visual sensor reading is updated to generate the reward signal for weight updates

## III. EXPERIMENTAL SETUP

### A. Reward Calculation

The reward combines both visual and gyro sensor inputs. The gyro sensor reward is partitioned into three categories as shown in Fig. 4. In the first case the robot moves less than three legs losing its balance and not moving forward. This results in a small positive reward to encourage more neurons to fire in the next time instance. The other case is when four or more neurons in the CPG fire making the corresponding legs to be raised, which results the robot losing balance and gives a small negative reward to make sure that the weights are depressed and fewer number of neurons fire in the next iteration. The final case is when the correct three legs are raised, and the robot maintains its balance even when the legs are lifted up in the air. A high positive reward is given now to make sure the action is repeated.

The other component of the reward comes from the visual input. We use the scan line based odometry method from [13] because of its low computational cost. If the balance is maintained because no leg is raised, it gives an erroneous positive reward to the system. This error can be fixed by using a binary decision algorithm taking difference between the images before and after the movement to distinguish whether the forward movement has occurred or not. A positive reward accompanies the forward movement and lack of any movement is penalized. This allows the robot to simultaneously learn how to balance itself while moving forward. It is important to ensure that the negative reward is not very high in magnitude, particularly in the first few iterations so that the network has enough time to explore and learn. A linear correction is applied to the movement-based reward making it steeper as the algorithm progresses. Therefore, the total reward at time instance t is given by

$$Reward_{total}[t] = Reward_{gyro} + Reward_{visual}\left[\frac{t}{T_1}\right] \quad (2)$$

**Algorithm 2** Gyroscope Reward Calculation
1: **for** $time = 1$ to $T$ **do**
2:   $G_{initial}$ = Read gyro sensor
3:   Complete the movement
4:   $G_{final}$ = Read gyro sensor
5:   **if** $(G_{final} - G_{initial} > G_{Thresh})$ **then**
6:     balance lost
7:     **if** (Number of legs moved $> 3$) **then**
8:       $Reward_{Gyro} = -2$
9:     **end if**
10:     **if** (Number of legs moved $< 3$) **then**
11:       $Reward_{Gyro} = +2$
12:     **end if**
13:   **end if**
14:   **if** $(G_{final} - G_{initial} < G_{Thresh})$ **then**
15:     balance maintained
16:     $Reward_{Gyro} = +5$
17:   **end if**
18: **end for**

Fig. 4. Pseudo-code for calculating the reward corresponding to gyro sensor reading. If the balance is maintained high positive reward is given. If the balance is lost, the reward is given such that the cumulative weights evolve 3 legs at a time

### B. Synaptic Weight Updating

Combining synaptic reinforcement with reward function has been utilized previously [14] speculating dopamine release as the biological parallel of reward signals. The weights are updated only for synapses connected to the CPG neurons firing at the current time instance. If neuron $i$ spikes at time step $t$, then for all pre-synaptic neurons spiking at time step $t$-1, the change of weight is calculated as given below

$$W_{ij}[t+1] = W_{ij}[t] + \theta\eta Reward_{total}[t] \quad (3)$$

where $\theta$ is the learning rate and $\eta = random(0,1)$ is the stochastic term. The learning rate has to be chosen carefully to maximize the convergence rate of learning process.

### C. Hardware Development

The system described above is implemented in hardware for verification of the correct gait pattern in an office environment shown in Fig. 1. We use an Adeept RaspClaws Hexapod Robot as the locomotion platform. It is equipped with a Gy-521 MPU-6050 MPU6050 gyro sensor Module and a pi camera to provide balance and visual input to a Raspberry pi 3 Model B+ processing unit. The quad-core processor of Raspberry Pi operates at 1.5GHz frequency. The schematic of connections and block diagram of the operations is shown in Fig. 5.

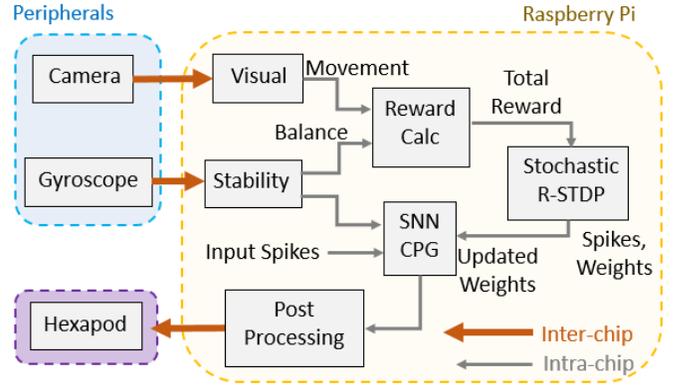

Fig. 5. Block diagram of the system of computation. Sensory inputs are processed on the Raspberry Pi board and the movement signals are generated and send out to the hexapod.

## IV. RESULTS AND DISCUSSION

### A. System Simulation

Fig. 6 shows the simulation results of an example of learning process along the time. The robot starts its action with no movement and gradually begin to oscillate between moving 2 and 4 legs alternatively. When it finally enters the target gait (the tripod gait shown in Fig. 1), thereby earning a high reward, and the weights saturate thus sustaining the gait. Fig. 6(b) shows the total accumulated reward over time. The reward was initially negative before the correct gait is found. After that, it rapidly increases.

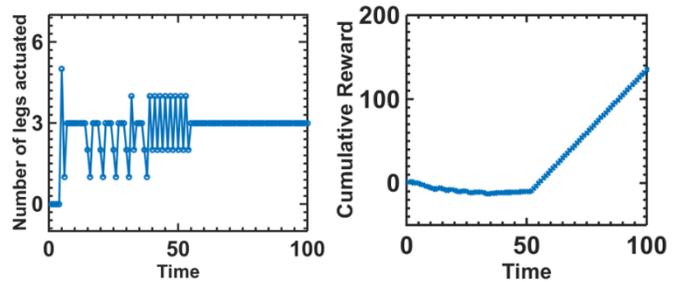

Fig. 6. The number of moving legs (a) and reward accumulated (b) during the learning process.

*B. Demonstration*

We apply the proposed method to an Adeept hexpod robot with the hardware configuration depicted in section III.C. In the first experimental demonstration (demo-1)[1], the learning process converged to the target gait pattern at the 66th cycle (4:10). We also note (demo 2) that the learning process converges to an unwanted alternative gait pattern at the 7th cycle (0:18). This gait pattern is a local minimum during the learning process. The hexapod can move also forward with this gait, but it is less efficient when compared to the bio-mimetic target gait. The occasional tremor of servos is caused by instantaneous drops in the current supply.

*C. Parameter Selection*

In order to identify the best learning rate, we simulated 100 iterations of the algorithm with different random seeds for different learning rates. This is shown in Fig. 7. Fig. 7(a) shows the percentage of simulations converging to the correct tripod gait. The results show that the convergence percentage remains similar over a range of learning rates. We choose 0.1 as the learning rate as it shows the lowest number of non-convergent cases. The non-convergent cases do not mean that no locomotion takes place. These cases correspond to an inefficient final gait, where the hexapod makes frequent turns and does not move forward in every step. Fig. 7(b) shows the convergence time for the iterations that converge to the desired gait. The median convergence time is 207 iterations for learning rate of 0.1. Fig. 7(c) shows the weight maps of synapses forming the connections between the CPG neurons before and after the completion of the learning. For the convergence to the global minima of tripod gait, the initially random weights between the neurons evolve into a pattern such that neurons 1,3,5 drive neurons 2,4,6 and vice versa.

*D. Comparison with the Prior Work*

The simple two-layer spiking network used in this network is expected to show high reduction in energy required in learning as opposed to conventional approaches involving artificial neural networks. We calculate the total number of spikes issued by the CPG in the course of learning to estimate the energy spent in learning. The median of total number of spikes issued is 503. If these dynamics were implemented in a specialized SNN processor, such as Intel's Loihi platform (1.7 nJ/spike [14]), then the total energy consumption in the learning process will be 855.1 nJ. Along with the low energy consumption, this work also shows that online SNN training can enable a hexapod to learn how to walk by a simple reward mechanism. (Table. 1)

## V. CONCLUSION

We design a closed-loop end-to-end online training of central pattern generator based on a bio-plausible spiking neural network. We have applied the proposed neuromorphic CPG to a hexapod robot and achieve autonomous online reinforcement learning of biomimetic walking gaits in an energy efficient manner. This online learning system is implemented on a resource-constrained embedded system. The learning process converges to the desired bio-observed tripod in 70% of the cases while in other cases it converges to suboptimal gaits that can still enable the locomotion.

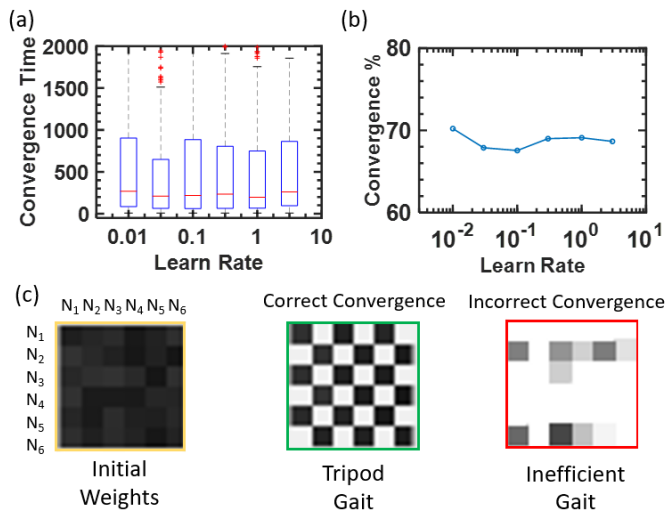

Fig. 7(a) Distribution of convergence time with learning rate. Median convergence time is 207-time instances. (b) Percentage of simulations converging to bio-observed tripod gait is independent of the learning rate. Simulations not converging to tripod gait are trapped in local minima that still enabling the locomotion is shown in demo 2. (c) Weight matrices of the CPG connections before and after the training. With the convergence to the global minima exhibiting tripod gait as shown in video demo - 1. In this configuration the neurons form a group of 3 reinforcing each other. In the other case the weights evolve into a configuration with inefficient walking gait still ensuring locomotion as shown in demo - 2

Table. I Benchmarking Comparison to previous SNN CPG approaches

| Ref | Training Approach | Sensory Feedback | Online/Offline |
|---|---|---|---|
| [16] | Linear Programming | None | Offline |
| [17] | Grammar Evolution | None | Offline |
| [18] | Reward + STDP | Olfactory + visual | Offline |
| This Work | Stochastic RL | Balance + visual | Online |

ACKNOWLEDGEMENT

This work was supported by C-BRIC, one of six centers in JUMP, a Semiconductor Research Corporation (SRC) program sponsored by DARPA.

---

Demo-1: https://youtu.be/1HqeISAkAs4
Demo-2: https://youtu.be/ypW0V23gEj0